\documentclass[11pt,oneside]{article}

\usepackage[a4paper,
            bindingoffset=0.2in,
            left=1in,
            right=1in,
            top=1in,
            bottom=1in,
            footskip=.25in]{geometry}

\usepackage{authblk}
\usepackage{epsfig}
\usepackage{graphicx}
\usepackage{subcaption}
\usepackage{calc}
\usepackage{amssymb}
\usepackage{amstext}
\usepackage{amsmath}
\usepackage{hyphenat}
\usepackage{amsthm}
\usepackage{multicol}
\usepackage{apalike}
\usepackage{booktabs}
\usepackage{framed,multirow}
\usepackage[table]{xcolor}
\usepackage{xspace}
\usepackage{soul}
\usepackage{color}
\let\bibhang\relax
\usepackage[numbers,sort&compress]{natbib}
\usepackage{url}
\usepackage{makecell}

\title{Proper Body Landmark Subset Enables More Accurate and 5X Faster Recognition of Isolated Signs in LIBRAS}

\author[1]{Daniele L. V. dos Santos}
\author[1]{Thiago B. Pereira}
\author{Carlos Eduardo G. R. Alves}
\author{Richard J. M. G. Tello}
\author{Francisco de A. Boldt}
\author{Thiago M. Paixão}
\affil[1]{Federal Institute of Espírito Santo, Campus Serra}
\affil[ ]{\normalsize \{danieleleite.vs,thiagoborges980,cadu97\}@gmail.com, \{richard,franciscoa,thiago.paixao\}@ifes.edu.br}
\date{}

\begin{document}
\maketitle

\begingroup
\renewcommand\thefootnote{}
\footnotetext{\textit{This work was accepted for presentation at IEEE SAS 2026.}}
\endgroup


\begin{abstract}
This paper examines the feasibility of utilizing lightweight body landmark detection for recognizing isolated signs in Brazilian Sign Language (LIBRAS). Although the use of skeleton-image representation has enabled substantial improvements in recognition performance, the use of OpenPose for landmark extraction hindered time performance. In a preliminary investigation, we observed that simply replacing OpenPose with lightweight MediaPipe, while improving processing speed, significantly reduced accuracy. To overcome this limitation, we explored landmark subset selection strategies to optimize recognition performance. Experimental results show that a proper landmark subset achieves comparable or superior performance to state-of-the-art methods while reducing processing time by more than 5$\times$. As an additional contribution, we demonstrate that spline-based imputation effectively mitigates missing landmark issues, leading to substantial accuracy gains.
\end{abstract}


\section{Introduction}

\label{sec:introduction}

Artificial intelligence has enabled significant advancements in accessible technologies across various domains, in particular, to support individuals with partial or total hearing loss: a population estimated at \(1.57\) billion in 2019, and projected to reach \(2.45\) billion worldwide by 2050 \cite{guo2024}. The literature has addressed sentence-level (continuous) \cite{hu2023continuous} and word-level (isolated) \cite{alves2024enhancing} recognition of Sign Language (SL) from video sequences. In the latter, the focus of this work, the goal is to identify a single sign (``word'') performed by an individual (signer) recorded in front of a camera. This application is particularly helpful for self-paced SL training platforms \cite{starner2024popsign}, visual keyword search for SL \cite{tamer2020keyword}, dictionary lookup by example \cite{bohacek2023sign}, as well as query-by-example sign spotting to locate occurrences in continuous videos \cite{varol2022scaling}, and cross-modal retrieval to search videos from free-form text \cite{duarte2022sign}.

Most works on Isolated Sign Language Recognition (ISLR) leverage deep neural networks, such as 2- and 3-D Convolutional Neural Networks (CNNs) \cite{alves2024enhancing,de2023automatic}, Gated Recurrent Units (GRUs) \cite{shen2024stepnet}, Long Short-term Memory (LSTM) networks \cite{rastgoo2021hand}, and Transformers \cite{erturk2025tslformer}. Promising results have been achieved by combining multiple modalities. For instance, De Castro \cite{de2023automatic} proposed a complex multi-stream 3-D CNN architecture in which each branch processed: joint distance and velocity maps; segmented hands; segmented faces; and the raw RGB frames augmented with a depth map. Despite its effectiveness, the approach introduces significant complexity and computational overhead, as landmark extraction depends on OpenPose \cite{cao2019openpose}, a framework that delivers good accuracy but at a high computational cost.

Conversely, Alves et al. \cite{alves2024enhancing} proposed a simpler, yet effective, single-stream pipeline. Body landmarks (trunk, hands, and face) are extracted from RGB frames and transformed into a 2-D skeleton-image representation \cite{memmesheimer2022skeleton}, which is fed into a CNN classifier. They achieved state-of-the-art (SOTA) accuracy on the two most popular LIBRAS ISLR datasets (\textsc{MINDS-Libras} \cite{rezende2021development} and \textsc{LIBRAS-UFOP} \cite{cerna2021multimodal}), however, at the cost of obtaining landmarks with OpenPose. To enhance efficiency and scalability, lightweight frameworks, such as MediaPipe \cite{lugaresi2019mediapipe}, have been explored. In \cite{erturk2025tslformer,luna2023interpreting}, the landmarks obtained with MediaPipe are directly processed by a Transformer network \cite{vaswani2017attention}, while in \cite{laines2023isolated}, the MediaPipe holistic landmarks (hands, face, and upper body) are converted into Tree-Structure Skeleton Images, later processed by a 2-D CNN.

The time efficiency enabled by MediaPipe has motivated us to investigate its impact on the ISLR framework proposed by Alves et al. \cite{alves2024enhancing}. However, preliminary exploration showed that using all landmarks provided by MediaPipe dramatically reduced recognition accuracy. Based on this observation, we investigate whether landmark subsets used in the literature, as well as in winning solutions publicly available from SLR challenges, enable the use of MediaPipe for \textsc{MINDS-Libras} and \textsc{LIBRAS-UFOP}. The main findings in this paper are:

\begin{itemize}
  \item A proper subset of landmarks substantially improves recognition performance while reducing input dimensionality.

  \item Simple spline-based imputation for missing landmarks further enhances accuracy.

  \item The best-performing landmark subset outperformed the literature on \textsc{MINDS-Libras} and \textsc{LIBRAS-UFOP}.

  \item Leveraging MediaPipe yields a speed-up of $5\times$ compared to the OpenPose counterpart.
\end{itemize}

\begin{figure}[t]
  \centering
  \includegraphics[width=0.8\textwidth]{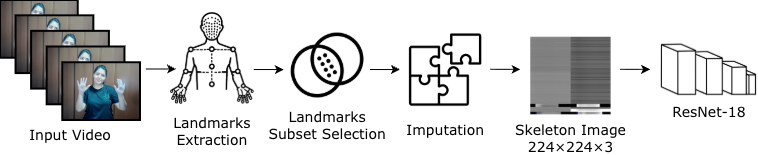}
  \caption{Overview of the ISLR framework. MediaPipe is used to detect landmarks, which are selected, interpolated, encoded as 2-D skeleton images, and then classified by a CNN.}
  \label{fig:pipeline}
\end{figure}
\section{Landmark-based ISLR Framework}
\label{sec:method}

The ISLR framework is illustrated in Figure \ref{fig:pipeline}. The input is a video sequence comprising a single sign. Body landmarks are extracted from each RGB frame using the MediaPipe Holistic model \cite{lugaresi2019mediapipe}, which combines pose, face, and hand markers into a unified full-body estimation. The landmarks are organized into a CSV file, where an even/odd column represents the $x$/$y$ normalized coordinate of a particular landmark, whereas the rows represent the temporal sequence. A subset of the detected landmarks is selected according to a predefined strategy. To impute missing points resulting from noisy detection, we apply spline-based interpolation, modeling temporally continuous trajectories for each point. The processed landmarks are then converted into a 2-D skeleton image representation \cite{memmesheimer2022skeleton}, which encodes the spatial structure and temporal dynamics of the landmarks. Finally, the image is resized to \(224\times224\) and fed into a 2-D CNN, which outputs the highest-probability sign.

The following sections detail the landmark extraction, landmark subset selection, imputation, image encoding, and the training process to deploy the classification model.


\subsection{Landmarks Extraction}

\begin{figure*}
  \centering
  \includegraphics[width=0.95\textwidth]{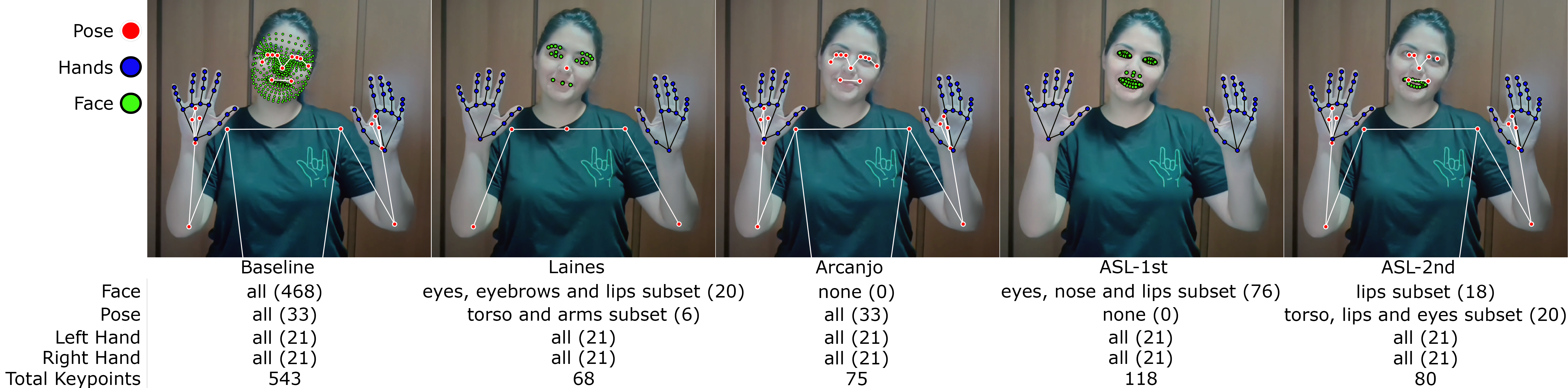}
  \caption{Landmarks subset selection: the columns show the set of landmarks utilized in each strategy and the respective count across the body parts.}
  \label{fig:strategies}
\end{figure*}

MediaPipe has been widely used in real-time human perception tasks due to its ability to capture manual articulators (hands), upper-body kinematics, and orofacial cues. Our ISLR pipeline relies on the Holistic model, which outputs full-body landmarks, including (by default) 543 points (Figure \ref{fig:strategies} (baseline)): 468 for the face, 33 for the pose (estimating body posture per frame across the entire body), and 21 per hand (42 in total). Note that the pose landmarks include a few coarse points around the head (e.g., nose and eyes), but the detailed facial landmarks are provided as a separate component. MediaPipe also provides depth information ($z$-coordinate) and visibility scores. Our approach, however, operates strictly in 2-D and therefore discards the additional information provided by MediaPipe.\\

\noindent \textbf{Settings.} The minimum detection confidence and minimum tracking thresholds in MediaPipe were set to 0.4, a value proven effective in preliminary tests for stabilizing landmark extraction. The former triggers the initial detection of the articulators (e.g., hands, face, and pose) in each frame, while the latter controls temporal tracking to maintain consistent trajectories across frames.

\subsection{Landmarks Subset Selection}
\label{sec:strategies}

As mentioned, 468 out of the 543 landmarks provided by the Holistic model correspond to the face alone, potentially introducing redundancy (non-linguistic variation) and noise. Based on the premise that appropriate landmark subsets are key to improve recognition, five strategies for landmark selection are considered in this study: the full set of points (baseline) and four subsets focusing on the most relevant articulators for SLR.

\subsubsection{Baseline (All)}
\label{subsec:strategy-all}

The baseline (Figure \ref{fig:strategies}, col. 1) employs all 543 landmarks provided by the Holistic model, serving as a reference for comparison against the other strategies. By retaining the full configuration, it maximizes spatial and articulatory information, preserving facial micro-configurations (mouth, eyebrows, and gaze), upper-limb kinematics, and manual gestures. The trade-off is a substantially larger input data and increased susceptibility to noise.

\subsubsection{Laines \cite{laines2023isolated}}
\label{subsec:strategy-laines}

This strategy (Figure \ref{fig:strategies}, col. 2) uses 68 landmarks distributed across the face, pose, and hands, prioritizing regions functionally relevant for signing (lips/expressivity, shoulder girdle and upper-limb joints, plus key points on both hands). The goal is to reduce dimensionality while preserving the articulatory mechanisms most critical for ISLR: manual configurations, arm/forearm orientation and perspective, and facial cues related to the mouth and expressiveness. Compared to the baseline (All), this strategy is expected to be more robust to noise and to benefit from signs strongly driven by the hands and salient facial/head regions.

\subsubsection{Arcanjo \cite{arcanjo2024}}
\label{subsec:strategy-arcanjo}

This configuration (Figure \ref{fig:strategies}, col. 3) retains pose and hands while entirely excluding the dense points of the face. It uses 33 pose points and 42 hand points, totaling 75 landmarks. As seen in Figure \ref{fig:strategies}, there are coarse points on the face provided by the pose component of MediaPipe. Removing the dense points from the face results in a substantial reduction in dimensionality, simplifying the hypothesis space and aiding regularization.

\subsubsection{ASL Signs Challenge 1st-place (ASL-1st)}
\label{subsec:strategy-d}

This strategy (Figure \ref{fig:strategies}, col. 4) adopts the subset from the winning solution \cite{sohn2023firstplace} of the Google ASL Signs Challenge on Kaggle (focus on isolated signs). A total of 118 landmarks were selected from the face and hands, excluding body pose. The emphasis is on manual configurations and the orofacial region (lips/jaw), which are distinctive in many lexicons, while global pose is discarded to reduce dimensionality and dependence on posture. The underlying observation is that, for many isolated signs, the differential cues lie in hand shape, contact/relative position to the face, and lip articulation rather than in trunk alignment.

\subsubsection{ASL Signs Challenge 2nd-place (ASL-2nd)}
\label{subsec:strategy-e}

Also derived from the Google ASL Signs Challenge on Kaggle, this configuration (Figure \ref{fig:strategies}, col. 5) replicates the subset from the runner-up solution \cite{peaceteam2023secondplace}, comprising 80 keypoints concentrated on the lips, hands, and body pose. Unlike ASL-1st, it preserves a core of pose landmarks to retain postural context (shoulder, elbow, and wrist angles) while still prioritizing manual articulation and the labial region. The aim is to balance fine-grained hand discrimination with sufficient global information to disambiguate signs with distinct trajectories or ranges.

\subsection{Spline-Based Landmark Imputation}
\label{sec:imputation}

Imputation is achieved through \textit{piecewise spline interpolation}, where each landmark-coordinate time series is treated as an independent function to be reconstructed. Missing values are estimated from their short temporal neighborhoods (window size of 5), leveraging information from both past and future frames. Specifically, we employ a cubic spline for robust reconstruction when at least four points are available in the series; otherwise, the procedure defaults to linear interpolation. The short-window constraint preserves kinematic plausibility and prevents extrapolation beyond the observed range.

\subsection{Image Encoding via Skeleton-DML}
\label{sec:encoding}

Landmarks are encoded as images using Skeleton-DML \cite{memmesheimer2022skeleton}, originally applied to represent skeletal joints in few-shot action recognition. To the best of our knowledge, Alves et al. \cite{alves2024enhancing} were the first to apply Skeleton-DML to ISLR, obtaining promising results on LIBRAS datasets. The underlying idea of such representation applies broadly to space–time coordinates, such as general body landmarks. Adapting it to the 2-D case is straightforward by disregarding depth ($z$-coordinate), which simply reduces the encoded data volume.

Given $L$ landmarks and $T$ frames, the 2-D version of the Skeleton-DML encoding procedure consists of creating two $L \times T$ matrices (tensors), one for each of the $x$ and $y$ coordinates. These tensors are reshaped so that three consecutive column entries are organized as channels, forming a 3-channel image. The reshaped tensors are then horizontally concatenated, forming a single image. More details are provided in the original work \cite{memmesheimer2022skeleton} and in \cite{alves2024enhancing}.

\subsection{Training Protocol}
\label{sec:training}

The training protocol follows \cite{alves2024enhancing}, except for the increasing in the number of epochs (from 20 to 30). At every epoch, we perform on-the-fly data augmentation directly on the landmark sequences prior to their encoding into 2-D images, consisting of rotation, zoom, translation, and horizontal flip to increase data diversity without changing the number of samples. The classifier is a ResNet-18 initialized with ImageNet weights to ensure a setup comparable to \cite{alves2024enhancing}. Its final convolutional output is first passed through a global average pooling layer, producing a 512-dimensional feature vector. This vector is then fed into two fully connected layers: a 128-unit layer with ReLU activation, followed by an output layer matching the number of classes. The network input is \(224\times224\), while landmark-encoded images have size \(126 \times (2T/3)\), where $T$ is the number of frames of that particular input video. For the evaluated datasets, \(2T/3 < 224\), so resizing to \(224\times224\) results in upsampling without information loss.

Optimization leverages Adam to minimize cross-entropy, with batch size \(64\), learning rate \(10^{-4}\), and 30 epochs. Early stopping is triggered if the validation loss fails to decrease for five consecutive epochs. To prevent overfitting, we apply batch normalization after the 512-dimensional layer, dropout (with 50\% probability) after the 128-unit layer, and L2 regularization (weight decay of \(10^{-4}\)) on all trainable parameters. The model selected for evaluation is the checkpoint achieving the highest validation accuracy.
\section{Experimental Methodology}
\label{sec:experiments}

This section outlines the performance assessment, including the Brazilian Sign Language (LIBRAS) datasets, evaluation metrics, experiments, and the computational setup.

\subsection{Datasets}

Two popular ISLR LIBRAS datasets were considered in this investigation: \textsc{MINDS-Libras} \cite{rezende2021development} and \textsc{LIBRAS-UFOP} \cite{cerna2021multimodal}. In particular, these collections enable direct comparison with \cite{alves2024enhancing}, whose methodology motivated this study.

\subsubsection{\textsc{MINDS-Libras}}

It comprises 1,155 video recordings -- each representing a sign -- covering 20 signs selected according to phonological parameters (handshape, location, orientation, movement, and non-manual markers). The executions were performed by 12 signers varying in sex, age, and fluency in LIBRAS. Although clothing colors were not strictly standardized, most signers wore dark (black) attire, minimizing background cluttering. Data was collected in a controlled environment with a fixed chroma key background and sensors positioned to capture the movement of the upper limbs.

Two acquisition modalities were employed: an RGB camera (Canon EOS Rebel T5i, resolution 1920 $\times$ 1080) and an RGB-D sensor (Kinect v2, depth resolution 640 $\times$ 480), with only the RGB modality leveraged in this study. The typical recording distances were (approximately) 1.60m for the RGB-D sensor and 1.92m for the RGB camera.

\subsubsection{\textsc{LIBRAS-UFOP}}

This is a multimodal dataset created with a Microsoft Kinect v1 device that captures synchronized recordings in three modalities: RGB, depth (RGB-D), and 3D skeleton data. Similar to \textsc{MINDS-Libras}, only the RGB modality is explored in our experiments. The dataset comprises 56 signs, carefully selected, grouped, and organized into four categories to emphasize fine-grained phonological distinctions.

The recordings were performed by five signers in two different scenarios, under diverse illumination conditions, and without strict standardization of the distance between the participant and the sensor. The signs were executed at different speeds and considered both one- and two-handed executions, depending on the lexical nature of the sign. Recordings were captured at 30 FPS, with a resolution of 640 $\times$ 480, totalling 3,040 video sequences. The original recordings comprise between 8 and 16 repetitions (in sequence) of the same sign for a given signer. For consistency with  the ISLR framework, which requires a single sign execution per video, we use the same time cut points as \cite{alves2024enhancing}.

\subsection{Recognition Performance Metrics}

Performance was measured in terms of traditional metrics in the ISLR domain \cite{de2023automatic,alves2024enhancing}: accuracy, as well as macro-averaged precision, recall, and F1-score. Accuracy quantifies the proportion of correct predictions over the total number of instances and serves as a global indicator of performance. Precision measures the reliability of positive predictions, while recall (sensitivity) measures the coverage of actual positives. To balance these aspects, the F1-score -- the harmonic mean of precision and recall -- often provides a more informative summary than accuracy in the presence of class imbalance.

\subsection{Experiments}

The conducted experiments address four research questions:
\begin{itemize}
  \item \textbf{RQ1 (Main Question):} How does landmark selection affect recognition performance?
  \item \textbf{RQ2:} What is the impact of the spline-based imputation method employed in our approach?
  \item \textbf{RQ3:} How does our method, configured with the best-performing strategy, compare with existing approaches in the literature \cite{alves2024enhancing,de2023automatic,cerna2021multimodal,passos2021gait}?
  \item \textbf{RQ4:} What speed-up is provided by our MediaPipe-based solution relative to the OpenPose-based pipeline by Alves et al. \cite{alves2024enhancing}?
\end{itemize}

\subsubsection{Main Experiment (RQ1 + RQ2)}
\label{sec:main-experiment}

For each video, the full set of 543 landmarks was extracted with the MediaPipe Holistic pipeline. We retained the original raw data stream, containing the spatiotemporal trajectories of all points, as well as a processed stream obtained after imputation (Section \ref{sec:imputation}) to mitigate detection or tracking failures. Note that both streams comprise the full set of detected landmarks, serving to establish a comparison baseline. From each stream, four additional were derived by applying the selection strategies (Laines, Arcanjo, ASL-1st, and ASL-2nd), resulting in 10 streams, later encoded into $224 \times 224 \times 3$ skeleton images.

Finally, each encoded image set was subjected to train-test sessions (Section \ref{sec:training}). Following best practices \cite{alves2024enhancing,de2023automatic}, we adopted a nested \emph{Leave-One-Person-Out} (LOPO) protocol, where data from one signer is used for testing while the remaining $n-1$ signers' data is used for training and validation. In the nested approach, each of the $n-1$ signers is detached once for validation, while the others are used for training. In total, $n(n-1)$ train–validation–test sessions were conducted.

\subsubsection{Comparison with SOTA (RQ3)}
\label{sec:sota-comparison}
This evaluation focused on the ASL-2nd strategy, which yielded the highest F1-score on both \textsc{MINDS-Libras} and \textsc{LIBRAS-UFOP} (Table \ref{tab:main-exp}). Our results were compared to Alves et al. \cite{alves2024enhancing}, which motivated this work; De Castro et al. \cite{de2023automatic} (a multi-stream 3-D CNN combining RGB and artificial depth maps); Cerna et al. \cite{cerna2021multimodal} (a two-stream CNN combining RGB/depth and skeleton data); and Naz et al. \cite{naz2025} (a graph-based network for skeleton data). We also tested \cite{alves2024enhancing} with 30 epochs (rather than 20) for fairer comparability.


\subsubsection{Speed-up Evaluation (RQ4)}

This experiment compares the time performance of our approach -- adopting the ASL-2nd landmark selection strategy -- with \cite{alves2024enhancing}. While they rely on OpenPose \cite{cao2019openpose} to extract landmarks, our approach employs MediaPipe \cite{lugaresi2019mediapipe} for faster landmark extraction. The shortest, average-length, and longest videos of each dataset were processed five times for timing evaluation. In each run, 20 inference sessions with random data were executed to mitigate GPU warm-up effects. Landmark extraction, the most time-consuming procedure, was measured for both approaches. Model inference was measured only for \cite{alves2024enhancing} since the same CNN model is employed in both cases.




\subsection{Experimental Platform}

\textbf{Speed-up evaluation:} Intel Core i7-13700K (5.4 GHz), 32 GB DDR4, and NVIDIA GeForce RTX 4070 GPU. \textbf{Other experiments:} Intel Core i9-10900KF (3.70 GHz), 32 GB RAM, and NVIDIA GeForce RTX 3060 GPU. The software stack used Python 3.10 and PyTorch 2.0.1 (CUDA 12.1). Source code, checkpoints, and supplementary material are available at \url{https://github.com/danielelvs/islr-subset}.
\section{Results and Discussion}
\label{sec:results}

\subsection{Main Experiment}
\label{sec:main-exp}

Table \ref{tab:main-exp} shows a comparative view for the five landmark subsets considered in this work. All metrics, including F1-score, were obtained with imputation enabled; the last column quantifies the improvement yielded by imputation in percent points (e.g., F1-score for Laines increases by 4 p.p. through imputation). For both datasets, the baseline (All) yielded the worst results. For \textsc{MINDS-Libras}, the four landmark subsets performed very similarly, with small fluctuations in the average and standard deviation values (within $\pm$ 2 p.p.). A more pronounced contrast was observed for \textsc{LIBRAS-UFOP}: ASL-2nd achieved an F1-score of 0.91, 5 p.p. higher than ASL-1st and 4 p.p. higher than Laines \cite{laines2023isolated}. In summary, ASL-2nd and Arcanjo \cite{arcanjo2024} presented a more consistent F1-score.

\begin{table}[ht]
  \centering
  \caption{Comparing landmark subsets (main experiment).}
  \label{tab:main-exp}
  \resizebox{\columnwidth}{!}{
    \begin{tabular}{lccccc}
      \toprule
      \textbf{Land. subset} & \textbf{Accuracy} & \textbf{Precision} & \textbf{Recall} & \textbf{F1-score} & \makecell{\textbf{F1} \\ \textbf{imp.}} \\
      \midrule
      \multicolumn{6}{l}{\textsc{MINDS-Libras}}\\
      \midrule
      All & 0.70 (0.07) & 0.71 (0.06) & 0.69 (0.07) & 0.66 (0.07) & 6 \\
      Laines\cite{laines2023isolated} & \textbf{0.95 (0.04)} & 0.95 (0.05) & \textbf{0.95 (0.04)} & \textbf{0.94 (0.04)} & 4 \\
      Arcanjo \cite{arcanjo2024} & \textbf{0.95 (0.05)} & \textbf{0.96 (0.03)} & 0.94 (0.04) & \textbf{0.94 (0.05)} & 4 \\
      ASL-1st \cite{sohn2023firstplace} & \textbf{0.95 (0.05)} & 0.95 (0.05) & \textbf{0.95 (0.05)} & \textbf{0.94 (0.05)} & 6 \\
      ASL-2nd \cite{peaceteam2023secondplace} & 0.94 (0.04) & 0.95 (0.05) & \textbf{0.95 (0.04)} & \textbf{0.94 (0.05)} & 4 \\
      \midrule
      \multicolumn{6}{l}{\textsc{LIBRAS-UFOP}}\\
      \midrule
      All & 0.72 (0.04) & 0.74 (0.02) & 0.72 (0.04) & 0.69 (0.04) & 18 \\
      Laines \cite{laines2023isolated} & 0.88 (0.05) & 0.90 (0.04) & 0.88 (0.05) & 0.87 (0.05) & 17 \\
      Arcanjo \cite{arcanjo2024} & \textbf{0.91 (0.04)} & 0.92 (0.04) & \textbf{0.91 (0.04)} & 0.90 (0.04) & 5 \\
      ASL-1st \cite{sohn2023firstplace} & 0.87 (0.05) & 0.89 (0.04) & 0.87 (0.05) & 0.86 (0.05) & 15 \\
      ASL-2nd \cite{peaceteam2023secondplace} & \textbf{0.91 (0.04)} & \textbf{0.93 (0.03)} & \textbf{0.91 (0.04)} & \textbf{0.91 (0.04)} & 6 \\
      \bottomrule
      \multicolumn{6}{l}{{\raggedright \scriptsize Average values followed by the standard deviation in parentheses (when applicable).}}\\
      \multicolumn{6}{l}{{\raggedright \scriptsize The last column reports the improvement (p.p.) yielded by spline-based imputation.}}\\
      \multicolumn{6}{l}{{\raggedright \scriptsize Values in \textbf{bold} indicate the highest average metric value for a given dataset.}}\\
    \end{tabular}
  }
\end{table}

The reduction in input dimensionality (i.e., fewer landmarks) benefits generalization while preserving the essential articulatory cues for isolated sign recognition. Unlike the baseline (All), the other subsets focus primarily on pose and hand landmarks, with reduced or no reliance on dense facial meshes. This design helps mitigate noise caused by factors such as longer camera distances, illumination variability, and occasional occlusions that affect facial landmark quality.

Overall, imputation enhanced the performance of all subsets by at least 4 p.p. The differences were more pronounced for \textsc{LIBRAS-UFOP}, where improvements of over 15 p.p. were observed in three cases. The obtained results reinforce the need for post-processing the landmarks, as their detection with MediaPipe can be highly unstable.

\begin{table*}[t]
  \caption{Comparison with the SOTA.}
    \label{tab:results-comparative}
      \centering
        \resizebox{0.9\textwidth}{!}{
    \begin{tabular}{lccccccc}
      \toprule
      \textbf{Method} & \textbf{Year of pub.} & \textbf{Dataset} & \textbf{LOPO Eval.?} & \textbf{Accuracy} & \textbf{Precision} & \textbf{Recall} & \textbf{F1-score} \\
      \midrule
      \textbf{Ours (ASL-2nd)} & & \multirow{6}{*}{\textsc{MINDS-Libras}} & \checkmark & 0.94 (0.04) & \textbf{0.95 (0.05)} & \textbf{0.95 (0.04)} & \textbf{0.94 (0.05)} \\
      Alves et al. \cite{alves2024enhancing} (30 epochs) & 2024 & & \checkmark & 0.94 (0.04) & 0.94 (0.05) & 0.94 (0.04) & 0.93 (0.04) \\
      Alves et al. \cite{alves2024enhancing} & 2024 & & \checkmark & 0.93 (0.05) & 0.94 (0.05) & 0.93 (0.05) & 0.93 (0.05) \\
      Naz et al. \cite{naz2025} & 2025 & & & \textbf{0.97 (0.01)} & - & - & - \\
      De Castro et al. \cite{de2023automatic} & 2024 & & \checkmark & 0.91 (0.07) & - & - & 0.90 (0.08) \\
      Passos et al. \cite{passos2021gait} & 2021 & & & 0.85 (0.02) & - & - & - \\
      \midrule
      \textbf{Ours (ASL-2nd)} & & \multirow{7}{*}{\textsc{LIBRAS-UFOP}} & \checkmark & \textbf{0.91 (0.04)} & \textbf{0.93 (0.03)} & \textbf{0.91 (0.04)} & \textbf{0.91 (0.04)} \\
      Alves et al. \cite{alves2024enhancing} (30 epochs) & 2024 & & \checkmark & 0.82 (0.03) & 0.84 (0.03) & 0.82 (0.03) & 0.80 (0.04) \\
      Alves et al. \cite{alves2024enhancing} & 2024 & & \checkmark & 0.82 (0.04) & 0.83 (0.05) & 0.81 (0.05) & 0.80 (0.05) \\
      Naz et al. \cite{naz2025} & 2025 & & \checkmark & 0.89 (0.04) & - & - & - \\
      De Castro et al. \cite{de2023automatic} & 2024 & & \checkmark & 0.74 (0.04) & - & - & 0.71 (0.05) \\
      Passos et al. \cite{passos2021gait} & 2021 & & \checkmark & 0.65 (0.04) & - & - & - \\
      Cerna et al. \cite{cerna2021multimodal} & 2021 & & \checkmark & 0.74 (0.03) & - & - & - \\
      \bottomrule
      \multicolumn{8}{l}{{\raggedright \scriptsize Average values followed by the standard deviation in parentheses (when applicable). Values in \textbf{bold} indicate the highest average metric value for a given dataset}}\\
      \multicolumn{8}{l}{{\raggedright \scriptsize The `-'' marker indicates that the metric was not reported in the original publication.}}\\
    \end{tabular}
  }
\end{table*}

\subsection{Comparison with SOTA}
\label{subsec:results-sota}

Table \ref{tab:results-comparative} shows the results of the comparative evaluation with the literature. Results for \cite{cerna2021multimodal} are reported only for \textsc{LIBRAS-UFOP}. On \textsc{MINDS-Libras}, our accuracy was comparable to \cite{alves2024enhancing}, which slightly benefited from increasing training (30 epochs rather than 20). Although \cite{naz2025} achieved the highest accuracy, their evaluation relied on a less robust evaluation protocol than LOPO. For reference, De Castro et al. \cite{de2023automatic} observed an accuracy drop of nearly 40 p.p. when switching from traditional $k$-fold to LOPO, highlighting how strongly results depend on the evaluation protocol.

On \textsc{LIBRAS-UFOP}, we outperformed the competing methods in all metrics, including \cite{naz2025} (+2 p.p. in accuracy), which did not employ LOPO. Compared to \cite{alves2024enhancing}, we achieved significative higher performance across all metrics (+11 p.p. in F1-score). These results reveal that a proper subset of landmarks, combined with an imputation procedure, is highly beneficial even when using lighter landmark extraction.

\begin{table}[t]
    \centering
    \caption{Time performace analysis (Speed-up Evaluation).}
    \label{tab:results-performance}
    \setlength{\tabcolsep}{4pt}
    \resizebox{0.95\columnwidth}{!}{%
\begin{tabular}{lccccccc}
\toprule
\multirow{2}{*}{\makecell{\textbf{Video} \\ \textbf{instance}}} 
& \multirow{2}{*}{\textbf{\#Frames}} 
& \multirow{2}{*}{\makecell{\textbf{Inference} \\ \textbf{time (s)}}} 
& \multicolumn{2}{c}{\textbf{Land. time (s)}} 
& \multicolumn{2}{c}{\textbf{SPS}} 
& \multirow{2}{*}{\textbf{Speed-up}} \\
\cmidrule(lr){4-5} \cmidrule(lr){6-7}
& & 
& \textbf{Ours} & \textbf{\cite{alves2024enhancing}} 
& \textbf{Ours} & \textbf{\cite{alves2024enhancing}} 
& \\
\midrule
ML-1 & 71  & 0.46 & 2.27 & 14.14 & 0.37 & 0.07 & 5.36 \\
ML-2 & 139 & 0.89 & 4.44 & 27.69 & 0.19 & 0.03 & 5.36 \\
ML-3 & 233 & 1.49 & 7.44 & 46.41 & 0.11 & 0.02 & 5.36 \\
\midrule
LU-1 & 28  & 0.18 & 0.84 & 5.58  & 0.98 & 0.17 & 5.65 \\
LU-2 & 67  & 0.43 & 2.14 & 13.35 & 0.39 & 0.07 & 5.36 \\
LU-3 & 121 & 0.78 & 3.86 & 24.10 & 0.22 & 0.04 & 5.36 \\
\bottomrule
\multicolumn{8}{l}{{\raggedright \scriptsize ML: \textsc{MINDS-Libras}; LU: \textsc{LIBRAS-UFOP}; SPS: Signs per second.}}
\end{tabular}}
\end{table}
            
\subsection{Speed-up Evaluation}
\label{sec:speed-up}

Table \ref{tab:results-performance} presents the results of the speed-up evaluation, including: inference and landmark-extraction avg. time; Signs Per Second (SPS), computed as the inverse of the total time (sum of average inference and landmark times); Speed-up factor, given by the ratio between the SPS of our method and that of \cite{alves2024enhancing}. The higher the frame count (\#Frames), the more SPS decreases, indicating that longer inputs lead to higher response latency, which may affect user experience. This effect could be mitigated by processing only a subset of frames; however, this analysis is beyond the scope of this work.

The frames-per-second (FPS) rate can be estimated by multiplying the video length by SPS, yielding 26.32 and 4.86 FPS for our method and \cite{alves2024enhancing}, respectively. This corresponds to a speed-up greater than 5$\times$, reinforcing the substantial efficiency gain of our method. Although our pipeline achieved a frame rate close to 30 FPS, strict real-time performance depends on the deployment configuration, such as the hardware setup. Although this was not our focus, the obtained results reveal an advance toward real-time ISLR.

\section{Conclusion and Future Work}
\label{sec:conclusion}

This work investigated the feasibility of lightweight body landmark detection with MediaPipe for the recognition of isolated signs in LIBRAS.
Preliminary investigation of ISLR framework in \cite{alves2024enhancing} indicated that simply replacing OpenPose with the MediaPipe Holistic model, although improved time efficiency, dramatically hindered recognition performance.

To address this issue, we investigated four strategies for selecting landmark subsets to optimize recognition accuracy. Experimental results revealed that using the landmark subset from the second-best solution of the Google ASL Signs Challenge on Kaggle (ASL-2nd) achieved the most consistent results, being comparable (or even superior) to SOTA methods. Compared to \cite{alves2024enhancing}, our approach produced optimized solutions in terms of both accuracy and time performance, achieving a runtime reduction of more than 5$\times$. Additionally, we showed that spline-based imputation has a strong impact on recognition, highlighting the importance of landmark post-processing.


Future work will explore other sign languages, datasets, and lower-end hardware, allowing for a more comprehensive investigation. We will also investigate the impact of using pre-trained models (instead of training from scratch) and the effects of spatial upsampling on recognition performance. 
Finally, we envision extending our method to continuous (sentence level) SLR.


\section*{Acknowledgements}

This work was supported by FAPES/UnAC (No. 1068/2023, P 2023-SGLQ7) through Sistema UniversidadES, and by CAPES/FAPES (No. 132/2021, P 2021-2S6CD) under the PDPG (Graduate Development Program - Strategic Partnerships in the States).

\bibliographystyle{IEEEtran}
{\small
\bibliography{references}}



\end{document}